\documentclass{article}

\usepackage{microtype}
\usepackage{graphicx}
\usepackage{subcaption}
\usepackage{booktabs} 

\usepackage[preprint]{icml2026}

\usepackage{amsmath}
\usepackage{amssymb}
\usepackage{mathtools}
\usepackage{amsthm}

\usepackage{hyperref}
\usepackage{enumitem}
\usepackage{wrapfig}
\usepackage{tikz}
\usepackage{pifont}
\definecolor{myorange}{RGB}{80, 101, 142}
\definecolor{lightblue}{RGB}{91, 182, 220}
\definecolor{bblue}{RGB}{51,102,204}
\definecolor{myred}{RGB}{204,0,0} 
\hypersetup{
    colorlinks=true,
    linkcolor=red,
    filecolor=myorange,
    urlcolor=lightblue,
    citecolor=lightblue,
}

\usepackage{url}
\usepackage{xurl}
\usepackage{amsfonts}       
\usepackage{nicefrac}       
\usepackage{bm}
\usepackage[normalem]{ulem}
\useunder{\uline}{\ul}{}
\usepackage{color}    
\usepackage{caption}
\usepackage[table]{xcolor}
\usepackage{tcolorbox}

\usepackage[capitalize,noabbrev]{cleveref}

\theoremstyle{plain}

\theoremstyle{definition}

\theoremstyle{remark}

\usepackage[textsize=tiny]{todonotes}

\icmltitlerunning{Enhancing Multi-Modal LLMs Reasoning via Difficulty-Aware Group Normalization}

\begin{document}

\twocolumn[
  \icmltitle{Enhancing Multi-Modal LLMs Reasoning via \\ Difficulty-Aware Group Normalization}



  \icmlsetsymbol{equal}{*}

  \begin{icmlauthorlist}
    \icmlauthor{Jinghan Li}{ustc}
    \icmlauthor{Junfeng Fang}{nus}
    \icmlauthor{Jinda Lu}{ustc}
    \icmlauthor{Yuan Wang}{ustc} 
    \icmlauthor{Xiaoyan Guo}{ustc} \\
    \icmlauthor{Tianyu Zhang}{ustc}
    \icmlauthor{Xiang Wang}{ustc}
    \icmlauthor{Xiangnan He}{ustc}
  \end{icmlauthorlist}

  \icmlaffiliation{ustc}{University of Science and Technology of China}
  \icmlaffiliation{nus}{National University of Singapore}

  \icmlcorrespondingauthor{Junfeng Fang}{fangjf1997@gmail.com}
  \icmlcorrespondingauthor{Xiang Wang}{xiangwang1223@gmail.com}

  \icmlkeywords{Machine Learning, ICML}

  \vskip 0.3in
]



\printAffiliationsAndNotice{}  

\begin{abstract}
Reinforcement Learning with Verifiable Rewards (RLVR) and Group Relative Policy Optimization (GRPO) have significantly advanced the reasoning capabilities of large language models. 
Extending these methods to multimodal settings, however, faces a critical challenge: the instability of \textit{std}-based normalization, which is easily distorted by extreme samples with nearly positive or negative rewards. 
Unlike pure-text LLMs, multimodal models are particularly sensitive to such distortions, as both perceptual and reasoning errors influence their responses. 
To address this, we characterize each sample by its \textbf{difficulty}, defined through perceptual complexity (measured via visual entropy) and reasoning uncertainty (captured by model confidence).
Building on this characterization, we propose \textbf{\underline{d}iffic\underline{u}lty-awa\underline{r}e group normal\underline{i}z\underline{a}tio\underline{n} (Durian)}, which re-groups samples by difficulty levels and shares the \textit{std} within each group. 
Our approach preserves GRPO's intra-group distinctions while eliminating sensitivity to extreme cases, yielding significant performance gains across multiple multimodal reasoning benchmarks.
\end{abstract}
\section{Introduction}

\begin{figure}[ht]
  \vskip 0.2in
  \begin{center}
    \centerline{\includegraphics[width=\columnwidth]{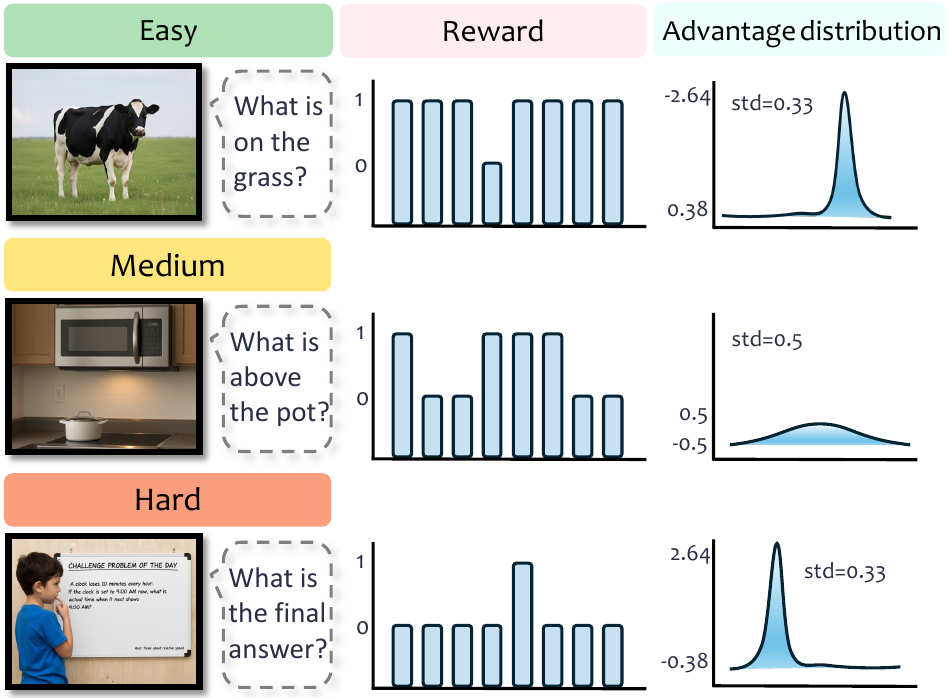}}
    \caption{
      The advantage distribution after the normalization of reward varies among samples. Extreme samples like easy and hard ones are amplified after \textit{std}-normalization, whereas medium samples exhibit more balanced advantages.}
    \label{fig:intro}
    \vskip -0.2in
  \end{center}
\end{figure}

Reinforcement Learning with Verifiable Rewards (RLVR) has enabled significant advances in the reasoning capabilities of both large language models (LLMs)~\citep{DeepSeekR1,qwen3,tulu3} and multi-modal large language models (MLLMs)~\citep{r1_reward, vision_r1}.
Within this paradigm, Group Relative Policy Optimization (GRPO)~\citep{DeepSeekMath} demonstrates strong performance by applying standard deviation (\textit{std})-based normalization to rewards within each response group.
This \textit{std}-based normalization rescales intra-group distinctions between positive and negative responses, thereby stabilizing training.

Despite these advances, we observe that the \textit{std}-based normalization suffers from a critical limitation: \emph{sensitive to extreme samples --- those with response groups that are almost entirely positive or negative}. 
Specifically, when rewards in a group collapse to near 0 or 1, the resulting low \textit{std} overemphasizes the extreme samples during optimization. Meanwhile, samples with more balanced rewards are neglected, leading to imbalanced optimization.
This issue is particularly pronounced in MLLMs, where the complexity of multimodal inputs increases the occurrence of such extreme samples. As illustrated in Figure \ref{fig:intro}, MLLM responses are jointly influenced by challenges from perceptual complexity and reasoning uncertainty, making them more susceptible to extreme reward distributions.

A straightforward solution is to remove the \textit{std} term thereby mitigating the risk of overfitting to extreme samples~\citep{dr.grpo}. However, it simultaneously discards the valuable intra-group distinctions, which are essential for effective and stable optimization.
Therefore, \emph{the key issue lies not in the std-normalization term itself, but rather in the way groups are constructed}: when group size is small, extreme cases become inevitable. Enlarging group sizes during rollouts could help, but it incurs prohibitive computational costs.

Motivated by this, we propose to account for the challenges of various samples, which we refer to as \textbf{Durian: difficulty-aware re-grouping}. We characterize each sample’s difficulty from two complementary perspectives: (i) a data-centric view, where the entropy of the image reflects \emph{perceptual difficulty}; and (ii) a model-centric view, where the confidence in model responses reflects \emph{reasoning difficulty}. 
By re-grouping samples according to these difficulty levels and sharing the \textit{std} within each group, our method preserves intra-group distinctions while mitigating sensitivity to extreme cases. Specifically, our difficulty-based re-group strategy is achieved by:

\textbf{Perceptual difficulty-based regrouping.} 
We quantify perceptual difficulty through spectral analysis of image patch covariances, where higher entropy in the resulting eigenvalue distribution indicates greater visual complexity~\citep{appendix_perceptual}. Images with more diverse and complex visual patterns exhibit higher entropy, reflecting greater perceptual difficulty.

\textbf{Reasoning difficulty-based regrouping.} 
Leveraging the insight that token-level log probabilities reflect reasoning confidence~\citep{rlpr}, we measure reasoning difficulty through the model's token-level confidence, where lower average log probabilities indicate greater uncertainty in generating correct reasoning chains, reflecting higher reasoning difficulty.

In summary, by explicitly decomposing difficulty into data-centric (\textbf{perceptual}) and model-centric (\textbf{reasoning}) groups, Durian allows each group of samples to share separate \textit{std}s for perceptual and reasoning aspects.
These normalized advantages are then combined to effectively integrate intrinsic data complexity and model uncertainty, ensuring stable optimization that preserves meaningful intra-group distinctions. To validate Durian, we conduct a comprehensive evaluation comparing it with leading methods on multiple benchmarks, and experimental results demonstrate that Durian attains more than 11.3\% average performance improvements.
\section{Preliminary}
\begin{figure*}[!t] 
    \centering  \includegraphics[width=0.9\textwidth]{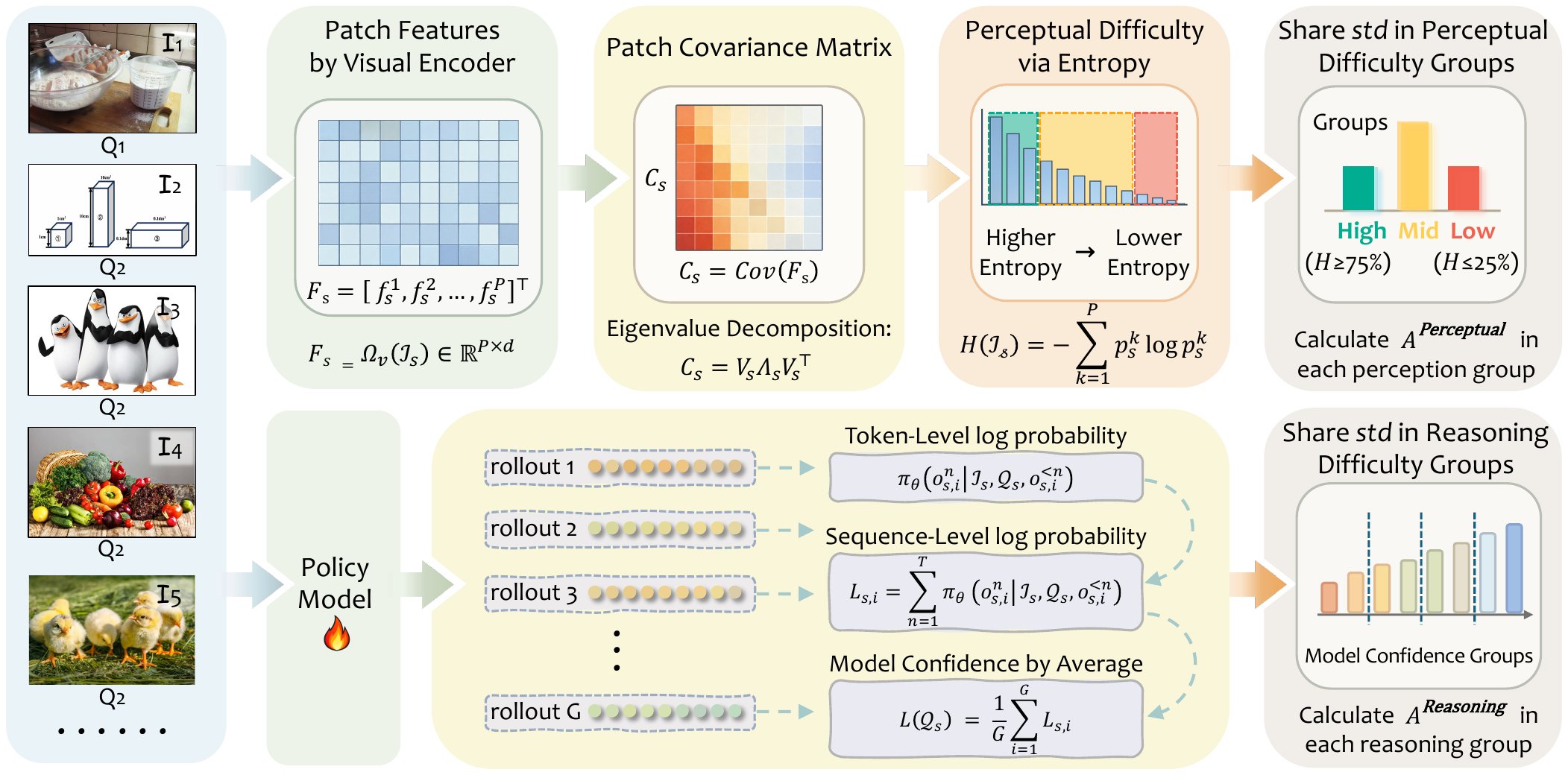}
    \caption{Overview of two difficulty-based regrouping strategies of Durian. Upper: For perceptual difficulty, we extract image patch features through the visual encoder and compute patch covariance matrices, whose eigenvalue entropy characterizes visual complexity. Bottom: For reasoning difficulty, model confidence is estimated from normalized sequence-level log probabilities across multiple rollouts. In both strategies, samples in the same group share the same \textit{std}.}
    \label{fig:method}
\end{figure*}
In this section, we introduce the key concepts and training setup for multimodal reasoning under RLVR \citep{DeepSeekR1}. We first formulate the task, and then revisit the standard GRPO framework \citep{DeepSeekMath} and its improved variant, Decoupled Clip and Dynamic Sampling Policy Optimization (DAPO) \citep{dapo}.

\subsection{Task Formulation}
We consider the problem of multimodal reasoning under the RLVR paradigm. Let $\{\mathcal{I,Q}\} \in \mathcal{D}$ denote a multimodal input, where the dataset $\mathcal{D}$ includes image $\mathcal{I}$ and text question $\mathcal{Q}$. 
The model generates a reasoning response $o$ given $\{\mathcal{I},\mathcal{Q}\}$ and receives a verifiable reward ${r}$ based on the correct answer $y$~\citep{synthrl,VLRethinker}. 
The response $o$ typically contains both the reasoning steps and the final answer, with the reasoning steps enclosed in \texttt{<think>...</think>} and the final answer enclosed in \texttt{\textbackslash boxed\{\}}.
We employ a binary reward function, where $r(o,y)=1$ if the final answer is equal to the correct answer $y$, and $r(o,y)=0$ otherwise.
The reasoning process is modeled as a policy $\pi_{\theta}(o|\mathcal{I,Q})$ parameterized by $\theta$ to maximize the expected reward:
\begin{equation}
\begin{aligned}
\mathcal{J}_{\mathrm{RLVR}}(\theta) = 
\displaystyle \max_{\theta} 
\mathbb{E}_{\{\mathcal{I,Q}\} \sim \mathcal{D}}
\mathbb{E}_{o \sim \pi_{\theta}(\cdot \mid \mathcal{I,Q})}[
r(o,y)].
\end{aligned}
\end{equation}
Our goal is to enhance the reasoning capabilities of an instruction-tuned MLLM, thereby significantly improving its performance on downstream multimodal reasoning tasks.

\subsection{Core Algorithms of Reinforcement Learning with Verifiable Reward}
\textbf{Group Relative Policy Optimization (GRPO)} is derived from Proximal Policy Optimization (PPO)~\citep{ppo}, with the key distinction that GRPO replaces the advantage estimates obtained via Generalized Advantage Estimation (GAE) with group-relative advantages computed from a group of outputs.

Specifically, for each input $\mathcal{I,Q}$, GRPO samples a group of outputs $\{o_1, o_2, \dots, o_G\}$ from the old policy model $\pi_{\theta_{\text{old}}}$, with rollout size $G$. The advantage of the $i$-th response is computed by normalizing the rewards among the group:
\begin{equation}
\begin{aligned}
\hat A_{i}
= \frac{r_i - mean(\{ r_1, r_2, \dots, r_G\})}{std(\{ r_1, r_2, \dots, r_G\})}.
\end{aligned}
\end{equation}

GRPO adopts a clipped objective, together with a directly imposed KL penalty term:
\begin{equation}
\resizebox{.91\hsize}{!}{$
\begin{aligned}
&\mathcal{J}_{\mathrm{GRPO}}(\theta) = 
\mathbb{E}_{(\mathcal{I,Q}) \sim \mathcal{D},\{o_i\}_{i=1}^G \sim {\pi_{\theta_{\text{old}}(o \mid \mathcal{I,Q})}}} \Bigg\{ \frac{1}{G} \sum_{i=1}^G \frac{1}{\left| o_i \right|}\sum_{t=1}^{\left| o_i \right|}\\
&\min\Bigg[ \frac{\pi_{\theta}\big(o_{i,t}\mid \mathcal{I,Q}, o_{i,<t}\big)}{\pi_{\theta_{\text{old}}}\big(o_{i,t}\mid \mathcal{I,Q}, o_{i,<t}\big)}\,\hat A_{i,t},\;
\mathrm{clip}\big(\frac{\pi_{\theta}\big(o_{i,t}\mid \mathcal{I,Q}, o_{i,<t}\big)}{\pi_{\theta_{\text{old}}}\big(o_{i,t}\mid \mathcal{I,Q}, o_{i,<t}\big)},\\
& 1-\epsilon,1+\epsilon\big)\,\hat A_{i,t}\Bigg] - \beta \mathbb{D}_{KL} (\pi_{\theta} \| \pi_{\text{ref}})
\Bigg\}. 
\end{aligned}$}
\end{equation}

$\epsilon$ is the hyperparameter to control the clipping range of the importance sampling ratio, and $\beta$ is the penalty strength of how far the current policy $\pi_\theta$ deviates from the reference policy $\pi_{ref}$.

\textbf{Decoupled Clip and Dynamic Sampling Policy Optimization (DAPO)} is a variant of GRPO adopting an asymmetric clipping range with a larger upper bound, dynamic sampling, token-level policy gradient loss, and overlong reward shaping. The objective function of DAPO is defined as:
\begin{equation}
\resizebox{.91\hsize}{!}{$
\begin{aligned}
&\mathcal{J}_{\mathrm{DAPO}}(\theta) = 
\mathbb{E}_{(\mathcal{I,Q}) \sim \mathcal{D},\,\{o_i\}_{i=1}^G\sim \pi_{\theta_{\text{old}}}(o \mid \mathcal{I,Q})} \Bigg\{\frac{1}{\sum_{i=1}^G \lvert o_i \rvert}  
\sum_{i=1}^G \sum_{t=1}^{\lvert o_i \rvert} \\
& \min \Bigg[\frac{\pi_{\theta}\big(o_{i,t}\mid \mathcal{I,Q}, o_{i,<t}\big)}{\pi_{\theta_{\text{old}}}\big(o_{i,t}\mid \mathcal{I,Q}, o_{i,<t}\big)}\,\hat A_{i,t},\;
\mathrm{clip}\bigg(\frac{\pi_{\theta}\big(o_{i,t}\mid \mathcal{I,Q}, o_{i,<t}\big)}{\pi_{\theta_{\text{old}}}\big(o_{i,t}\mid \mathcal{I,Q}, o_{i,<t}\big)}, \\
& 1-\epsilon_{\text{low}}, 1+\epsilon_{\text{high}}\bigg)\,\hat A_{i,t}
\Bigg]\Bigg\}. 
\end{aligned}$}
\end{equation}
\section{Durian\includegraphics[width=0.06\columnwidth]{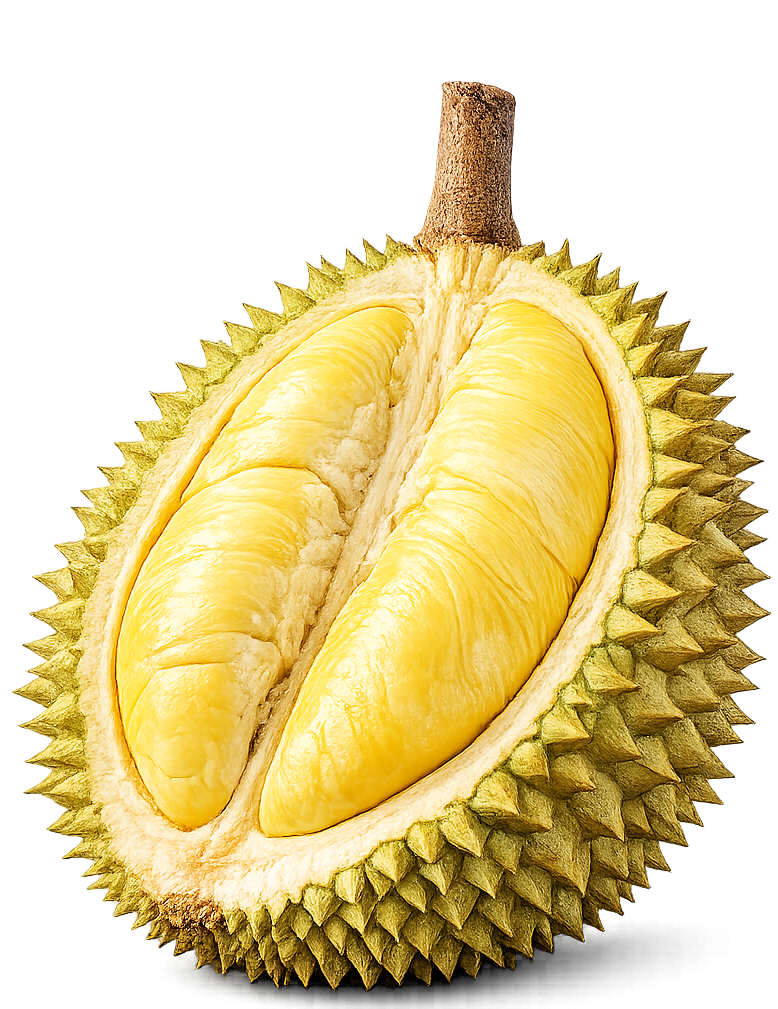}: Difficulty-based Regrouping }
In this section, we introduce our difficulty-based regrouping strategy in detail. 
We first represent our perceptual difficulty-based regrouping in Section \ref{sec:data-hardness}, then we describe our reasoning difficulty-based regrouping in Section \ref{sec:model-hardness}. The two regrouping strategies are summarized in Figure \ref{fig:method}.
Finally, we show the combination of these two strategies in Section \ref{sec:combine}.

\subsection{Perceptual Difficulty-based Regrpouping}
\label{sec:data-hardness}
\textbf{Perceptual difficulty estimation.}
To estimate the perceptual difficulty of a batch $\mathcal{B}=\{(\mathcal{I}_s, \mathcal{Q}_s)\}_{s=1}^{B}$, we first extract patch-level visual features from the Qwen2.5-VL-7B visual encoder $\mathbf{\Omega}_{v}$:
\begin{equation}
\begin{aligned}
\mathbf{F}_s = \mathbf{\Omega}_{v}(\mathcal{I}_s)\in\mathbb{R}^{P\times d}=[\boldsymbol{f}_{s}^{1}, \boldsymbol{f}_{s}^{2}, \cdots, \boldsymbol{f}_{s}^{P}]^\top,
\end{aligned}
\end{equation}
where $P$ denotes the number of spatial patches and $d$ is the feature dimension, with $\boldsymbol f_{s}^{j}\in \mathbb{R}^{d\times1}, j=1,\cdots, P $ representing the feature of the $j$-th patch.

Compared to CLIP-based representations~\citep{CLIP}, these patch-level features not only capture finer spatial granularity that preserves local details, but also align better with the downstream textual decoder $\mathbf{\Omega}_{t}$, ensuring both stability and semantic consistency.

We then compute the empirical covariance matrix to capture intra- and inter-patch variances:
\begin{equation}
\resizebox{.91\hsize}{!}{$
\begin{aligned}
\mathbf{C}_s 
= \tfrac{1}{P-1}\,(\mathbf{F}_s - \boldsymbol{1}_P\boldsymbol{\mu}_s^\top)
(\mathbf{F}_s - \boldsymbol{1}_P\boldsymbol{\mu}_s^\top)^\top,\quad
\boldsymbol{\mu}_s=\tfrac{1}{P}\sum_{j=1}^P \boldsymbol{f}_{s}^{j},
\end{aligned}$}
\end{equation}
where $\boldsymbol{1}_{P}$ is a $P\times 1$ column vector of ones and $\boldsymbol{\mu}_{s}$ is the mean of the patches feature of the $\mathcal{I}_s$.
The diagonal entries measure the variance of each feature dimension across patches, while the off-diagonal terms capture correlations between different feature dimensions. 
This covariance structure reveals whether visual features are dominated by a few strong dimensions or by multiple interacting factors, providing a principled basis for assessing perceptual difficulty.

Since $\mathbf{C}_s$ is a symmetric positive semidefinite matrix, we perform eigenvalue decomposition for spectral analysis:
\begin{equation}
\resizebox{.91\hsize}{!}{$
\begin{aligned}
\mathbf{C}_s = \mathbf{V}_s \mathbf{\Lambda}_s \mathbf{V}_s^\top,\quad 
\mathbf{\Lambda}_s=\mathrm{diag}(\lambda_{s}^{1},\ldots,\lambda_{s}^{P}),\;\lambda_{s}^{k}\ge 0.
\end{aligned}$}
\end{equation}
$\lambda_{s}^{k}$ denotes the $k$-th eigenvalue, quantifying the variance along one orthogonal principal direction. 
Concentrated eigenvalues indicate that most variance is captured by a few dimensions, whereas more balanced eigenvalues imply richer visual structure and higher visual complexity.

The final perceptual difficulty score is defined as the entropy~\citep{shannon_entropy} of the normalized distribution of eigenvalues:
\begin{equation}
\begin{aligned}
H(\mathcal{I}_s) = -\sum_{k=1}^{P} p_{s}^{k}\log p_{s}^{k},
\end{aligned}
\end{equation}
where $p_{s}$ is the normalized probability distribution of the eigenvalues, and each element in $p_{s}$ is calculated as:
\begin{equation}
\begin{aligned}
p_{s}^{k}=\tfrac{\lambda_{s}^{k}}{\sum_{j=1}^P \lambda_{s}^{j}},\quad \text{with} \sum_{k=1}^P p_{s}^{k}=1.
\end{aligned}
\end{equation}
Here, low entropy corresponds to the visually easy sample, with variance concentrated on a few dominant components, whereas high entropy indicates the visually difficult sample, with variance distributed across many patches.

\textbf{Perceptual difficulty-based regrouping.}
Given the perceptual difficulty scores within a batch, we partition samples into three groups using the 25th and 75th percentiles $\tau_{0.25}$ and $\tau_{0.75}$:
\begin{equation}    
\begin{cases}
\mathcal{S}_1=\{s \mid H(\mathcal{I}_s)\le\tau_{0.25}\},\quad \\
\mathcal{S}_2=\{s \mid\tau_{0.25}<H(\mathcal{I}_s)<\tau_{0.75}\},\quad \\
\mathcal{S}_3=\{s \mid H(\mathcal{I}_s)\ge\tau_{0.75}\}.
\end{cases}
\end{equation}


For each group $a$, the reward set can be defined as:
\begin{equation}
\begin{aligned}
    \mathcal{R}_a = \{r_{s, i}\mid i=1,\cdots,G ,s\in \mathcal{S}_a \}, \; a\in\{1,2,3\},
\end{aligned}
\end{equation}
where $r_{s, i}$ refers to the $i$-th reward of the $s$-th sample which belongs to the group $\mathcal{S}_a$.

We can then compute the shared standard deviation $std(\mathcal{R}_a)$ of group rewards, and normalize the reward of each sample in batch with the new $std(\mathcal{R}_a)$ to calculate advantage accordingly:
\begin{equation}
\begin{aligned}
A_{s, i}^{\text{Perceptual}}=\frac{r_{s,i}- mean(r_{s, 1}, r_{s, 2}, \dots r_{s, G})}{std(\mathcal{R}_a)\,}, 
\end{aligned}
\end{equation}
\begin{equation}
\resizebox{.9\hsize}{!}{$
\begin{aligned}
std(\mathcal{R}_a)=\sqrt{\frac{1}{|\mathcal{R}_a|-1}\sum_{r_{s, i}\in \mathcal{R}_a}\big(r_{s, i}-\frac{1}{|\mathcal{R}_a|}\sum_{r_{s, i}\in \mathcal{R}_a} r_{s,i}\big)^2}\,,
\end{aligned}$}
\end{equation}
where $|\mathcal{R}_a|$ denotes to the cardinality of $\mathcal{R}_a$.

By grouping samples into low-, medium-, and high-entropy categories, the normalization scale is shared only among samples with comparable perceptual difficulty. This mitigates the influence of extreme samples, balances treatment across different levels of visual complexity, and ultimately stabilizes optimization.

\subsection{Reasoning Difficulty-based Regrouping}
\label{sec:model-hardness}

\textbf{Reasoning difficulty estimation.}
While perceptual difficulty captures the intrinsic complexity of the image, reasoning difficulty is shaped by the model's intrinsic confidence in generating the final answer. 
Even for inputs with similar visual complexity, the model may exhibit varying confidence levels: high confidence (assigning a high probability to reasoning chains) implies a clear and reliable reasoning path, whereas low confidence indicates uncertainty and potential reasoning failures. 
Following this intuition, we quantify reasoning difficulty using the model's probabilities for its reasoning chains.

For the given batch $\mathcal{B}=\{(\mathcal{I}_s, \mathcal{Q}_s)\}_{s=1}^{B}$, and the generated $G$ responses for each sample, we denote the $i$-th response as $o_{s,i}=(o_{s,i}^1,\ldots,o_{s,i}^T)$, where $o_{s,i}^n$ is the $n$-th token and $T$ is the sequence length. 

Based on token-level log probability $\pi_\theta\!\;\big(o_{s,i}^n \,\big|\, \mathcal{I}_s,\mathcal{Q}_s, o_{s,i}^{<n}\big)$, we aggregate across tokens to obtain the sequence-level log probability for response $o_{s,i}$:
\begin{equation}
\begin{aligned}
L_{s,i}= \sum_{n=1}^{T} \pi_\theta\!\;\big(o_{s,i}^n \,\big|\, \mathcal{I}_s,\mathcal{Q}_s, o_{s,i}^{<n}\big).
\end{aligned}
\end{equation}
Then we define model confidence for sample $(\mathcal{I}_s, \mathcal{Q}_s)$ as the average sequence-level log probability across its $G$ rollouts:
\begin{equation}
\begin{aligned}
L(\mathcal{Q}_s) \;=\; \frac{1}{G}\sum_{i=1}^{G} L_{s,i}.
\end{aligned}
\end{equation}
This formulation reflects the model's internal confidence: High and consistent $L(\mathcal{Q}_s)$ indicates reliable reasoning chains, whereas low or fluctuating $L(\mathcal{Q}_s)$ reflects epistemic uncertainty, implying a more challenging reasoning sample.

\textbf{Reasoning difficulty-based regrouping.}
Given the model confidence scores $L(\mathcal{Q}_s)$ for the batch $\mathcal{B}=\{(\mathcal{I}_s, \mathcal{Q}_s)\}_{s=1}^{B}$, we divide samples into $b$ groups according to the quantiles of their confidence distribution. 
Let $\{\tau_0,\tau_1,\ldots,\tau_b\}$ denote the quantile boundaries, with $\tau_0 = 0$ and $\tau_b= 1$. 
Each question $\mathcal{Q}_s$ is then assigned to a group by: 
\begin{equation}
\resizebox{.89\hsize}{!}{$
\begin{aligned}
\mathcal{M}_u = \{\,s \mid \tau_{u-1} \leq L(\mathcal{Q}_s) < \tau_u\,\}, \quad u \in \{1,\ldots,b\}.
\end{aligned}$}
\end{equation}
Within each group $\mathcal{M}_u$, we define the reward set as:
\begin{equation}
\resizebox{.85\hsize}{!}{$
\begin{aligned}
    \mathcal{R}_u=\{ r_{s,i} \mid i=1,\dots,G, s \in \mathcal{M}_u \},\;u \in \{1,\dots, b\},
\end{aligned}$}
\end{equation}
where $r_{u,i}$ is the reward of the $i$-th response for the sample, which belongs to the $u$ group. We can then calculate the shared standard deviation $std(\mathcal{R_\mathcal{W}})$ of reasoning difficulty-based group, and compute the advantage accordingly:
\begin{equation}
\begin{aligned}
A_{s, i}^{\text{Reasoning}}=\frac{r_{s,i}- mean(r_{s, 1}, r_{s, 2}, \dots r_{s, G})}{std(\mathcal{R}_u)\,}, 
\end{aligned}
\end{equation}
where the \textit{std} can be calculated as:
\begin{equation}
\resizebox{.89\hsize}{!}{$
\begin{aligned}
std(\mathcal{R}_u)=\sqrt{\frac{1}{|\mathcal{R}_u|-1}\sum_{r_{s, i}\in \mathcal{R}_u}\big(r_{s, i}-\frac{1}{|\mathcal{R}_u|}\sum_{r_{s, i}\in \mathcal{R}_u} r_{s,i}\big)^2}\,,
\end{aligned}$}
\end{equation}
This regrouping ensures that responses with similar confidence levels are normalized on comparable scales, mitigating instability introduced by overconfident or underconfident samples.

\subsection{Combination for Robust Optimization}
\label{sec:combine}

To leverage the complementary aspects of perceptual and reasoning difficulty, we propose an element-wise combination strategy. 
Specifically, given the perceptual-based group normalized advantage $A^\text{Perceptual}$, the reasoning-based group normalized advantage $A^\text{Reasoning}$, and the original GRPO advantage $A^\text{GRPO}$, the combined advantage is defined as:
\begin{equation}
\resizebox{.52\hsize}{!}{$
\begin{aligned}
A^{\text{Combined}} \; & =\; \alpha_{\text{Ori}} \cdot A^{\text{GRPO}} \; \\ 
& +\; \alpha_{\text{Percep}} \cdot A^{\text{Perceptual}} \; \\
& +\; \alpha_{\text{Reason}} \cdot A^{\text{Reasoning}}, 
\end{aligned}$}
\end{equation}
where $\alpha_{\text{Ori}}, \alpha_{\text{Percep}}, \alpha_{\text{Reason}}$ are weighting coefficients that balance the contributions of the three components.
Perceptual difficulty, quantified by the entropy in the image, captures the \emph{visual complexity} of multimodal inputs; reasoning difficulty, derived from token- and sequence-level log probabilities, reflects the \emph{model uncertainty} during reasoning. 
Integrating these difficulty-based advantages with the original GRPO advantage allows the model to preserve meaningful intra-sample distinctions and incorporate both intrinsic and extrinsic difficulty context, providing a more stable and informative advantage for policy optimization.
\section{Experiment}
\label{sec:experiment}

\setlength{\tabcolsep}{3pt}
\begin{table*}[t]
    \caption{Performance comparison of Multi-modal LLMs with over 5 benchmarks. Accuracy scores (\%) are reported for all benchmarks for clarity. Data sizes used for SFT and RL are annotated in \textcolor{bblue}{blue} and \textcolor{myred}{red}, respectively. The best value in each column is shown in \textbf{bold}, and the second-best is \underline{underlined}.}
    \centering
    \resizebox{\linewidth}{!}{
        \begin{tabular}{lrcccccc}
            \toprule
            \textbf{Model} & \textbf{Data Size} & \textbf{MathVerse} & \textbf{MathVision} & \textbf{MathVista} & \textbf{WeMath} & \textbf{HallusionBench} & \textbf{Average} \\
            \midrule
            \multicolumn{6}{l}{\textit{Close-source models}} \\
            \midrule
            GPT-4o & - & 50.8 & 30.4 & 63.8 & 69.0 & 71.4 & -\\
            Claude-3.5-Sonnet & - & 26.5 & 38.0 & 67.7 & - & 71.6 \\
            \midrule
            \multicolumn{7}{l}{\textit{Open-source models}} \\
            \midrule
            InternVL-2.5-8B-Instruct ~\citep{internvl25} & - & 39.5 & 19.7 & 64.4 & - & 67.3 & - \\
            LLaVA-OneVision-7B ~\citep{llavaonevision} & - & 26.2 & - & 63.2 & - & 48.4 & - \\
            Kimi-VL-16B ~\citep{kimivl} & - & 44.9 & 21.4 & 68.7 & - & 66.2 & - \\
            URSA-8B ~\citep{ursa} & - & 45.7 & 26.2 & 59.8 & - & - & - \\
            Mulberry-7B ~\citep{mulberry} & - & - & - & 63.1 & - & - & - \\
            \midrule
            \multicolumn{7}{l}{\textit{reinforcement learning with verifiable reward based}} \\
            \midrule
            R1-VL-7B ~\citep{r1vl} & \textcolor{bblue}{260K}+\textcolor{myred}{10K} & 52.2 & 28.2 & 74.3 & 69.0 & 57.2 & 56.2 \\
            Vision-R1-7B ~\citep{visionr1} & \textcolor{bblue}{200K}+\textcolor{myred}{10K} & \underline{52.4} & 27.2 & \textbf{73.5} & 62.9 & 69.2 & 57.0\\
            R1-OneVision-7B ~\citep{r1onevision} & \textcolor{bblue}{155K}+\textcolor{myred}{10K} & 46.1 & 22.5 & 63.9 & 62.1 & 65.6 & 52.0\\
            OpenVLThinker-7B ~\citep{openvlthinker} & \textcolor{bblue}{35K}+\textcolor{myred}{15K} & 48.0 & 25.0 & 71.5 & 67.8 & 70.8 & 56.6\\
            MM-Eureka-Qwen-7B ~\citep{mmeureka} & \textcolor{myred}{15K} & 50.5 & 28.3 & 71.5 & 65.5 & 68.3 & 56.8\\
            ADORA-7B ~\citep{adora} & \textcolor{myred}{2.1K} & 50.1 & 27.6 & 71.1 & 67.1 & 53.1 & 53.8\\
            ThinkLite-7B-VL ~\citep{thinklite} & \textcolor{myred}{11K} & 50.2 & 27.6 & 72.7 & 69.2 & 71.0 & 58.1\\
            VLAA-Thinker-7B ~\citep{vlaa} & \textcolor{myred}{25K} & 49.9 & 26.9 & 68.8 & 67.9 & 68.6 & 56.4\\
            NoisyRollout~\citep{noisyrollout} &\textcolor{myred}{2.1K} & \textbf{53.2} & 28.5 & \underline{72.6} & \underline{69.6} & \underline{72.1} & \underline{59.2} \\
            
            \midrule
            Qwen2.5-VL-7B-Instruct ~\citep{qwen25} & - & 46.2 & 25.0 & 67.5 & 63.1 & 64.6 & 53.3\\
            + Vanilla GRPO & \textcolor{myred}{2.1K (Geometry3K)} & 49.6 & 26.8 & 70.2 & 68.2 & 69.8 & 56.9\\
            \rowcolor{blue!5}+ Durian (based on Vanilla GRPO) & \textcolor{myred}{2.1K (Geometry3K)} & \underline{52.8} & \underline{28.8} & 72.3 & 69.2 & \textbf{72.9} & \underline{59.2} \\
            + Vanilla DAPO & \textcolor{myred}{2.1K (Geometry3K)} & 50.4 & 27.6 & 70.7 & 69.4 & 68.6 & {57.3} \\
            \rowcolor{blue!5} + Durian (based on Vanilla DAPO) & \textcolor{myred}{2.1K (Geometry3K)} & 51.9 & \textbf{29.0} & 72.2 & \textbf{71.8} & 71.4 & \textbf{59.3}  \\
            \rowcolor{blue!5} + Durian (based on Vanilla DAPO) & \textcolor{myred}{39K (ViRL39K)} & 52.4 & 29.9 & 73.8 & 72.0 & 72.5 & 60.1 \\
            \bottomrule
        \end{tabular}
        }
    \label{tab:main_result}
\end{table*}

In this section, we introduce the key concepts and training setup for multimodal reasoning under RLVR \citep{DeepSeekR1}. We first formulate the task, and then revisit the standard GRPO framework \citep{DeepSeekMath} and its improved variant, Decoupled Clip and Dynamic Sampling Policy Optimization (DAPO) \citep{dapo}.

Specifically, we conduct comprehensive experiments to address the following research questions:

$\bullet$ RQ1: How does Durian perform on multimodal reasoning tasks compared to other baseline methods? \\
$\bullet$ RQ2: How do key components of Durian influence its performance?\\
$\bullet$ RQ3: How is the sensitivity of Durian under varying hyperparameters?

\subsection{Experimental Settings}
\textbf{Dataset.} For training, we rely on the Geometry3K ~\citep{geometry3k} dataset, which provides 2.1K training samples and 0.3K validation samples. Besides, we also provide experimental results training on a larger dataset ViRL39k.

\textbf{Benchmark.} We evaluate Durian on five benchmarks: four visual reasoning datasets, namely MathVerse ~\citep{mathverse}, MathVision ~\citep{mathvision}, MathVista ~\citep{mathvista}, and WeMath ~\citep{wemath}, as well as one visual perception benchmark, HallusionBench ~\citep{hallusionbench}. In addition, we assess the in-domain performance by comparing Durian with the vanilla GRPO and DAPO.

\textbf{Baseline.} To evaluate the performance of Durian, we consider three categories of baselines: 
(1) Closed-source models: GPT-4o~\citep{hurst2024gpt}, and Cloud-3.5-sonnet~\citep{Claude3.5}.
(2) Open source models: InternVL-2.5-8B-Instruct ~\citep{internvl25}, LLaVA-OneVision-7B ~\citep{llavaonevision}, Kimi-VL-16B ~\citep{kimivl}, URSA-8B ~\citep{ursa}, and Mulberry-7B ~\citep{mulberry}. 
(3) RLVR-based Models: MLLMs trained with reinforcement learning using verifiable rewards, representing the current mainstream approaches in this line of research. This category includes R1-VL-7B ~\citep{r1vl} , Vision-R1-7B ~\citep{visionr1} , R1-OneVision-7B ~\citep{r1onevision} , OpenVLThinker-7B ~\citep{openvlthinker} , MM-Eureka-Qwen-7B ~\citep{mmeureka} , ADORA-7B ~\citep{adora} , ThinkLite-7B-VL ~\citep{thinklite}, and VLAA-Thinker-7B ~\citep{vlaa}.

\textbf{Implementation details.} Following prior work ~\citep{noisyrollout}, we use Qwen2.5-VL-7B~\citep{qwen25} as base model and adpot EasyR1~\citep{easyr1} as reinforcement learning framework. All experiments are conducted on 8 NVIDIA H20 96G GPUs. We adopt the default settings from EasyR1, using a learning rate of $1e^{-6}$, a global batch size of 128, a rollout batch size of 512, and a rollout size of 8. The analysis of rollout size is provided in Appendix \ref{sec:rollout_size}.

\subsection{Comparison with Baseline Methods (RQ1)}
We comprehensively compare Durian with various state-of-the-art methods, and experimental results are listed in Table~\ref{tab:main_result}.
We can draw the following observations: (1) compared with those either distilled from large-scale chain-of-thought data or employing complex data augmentation strategies, our method, utilizing only 2.1k training samples, achieves comparable or even superior performance, significantly demonstrating our effectiveness.
(2) Building upon both GRPO and DAPO, our strategy demonstrates promising performance gains. Specifically, we achieve an average of 11.3\% improvements over Qwen2.5-VL, especially on the Mathvision, our strategy achieves more than 16\% improvements, further showing our effectiveness.

\subsection{Ablation Studies (RQ2)}
\begin{figure*}[!t] 
    \centering  \includegraphics[width=0.85\textwidth]{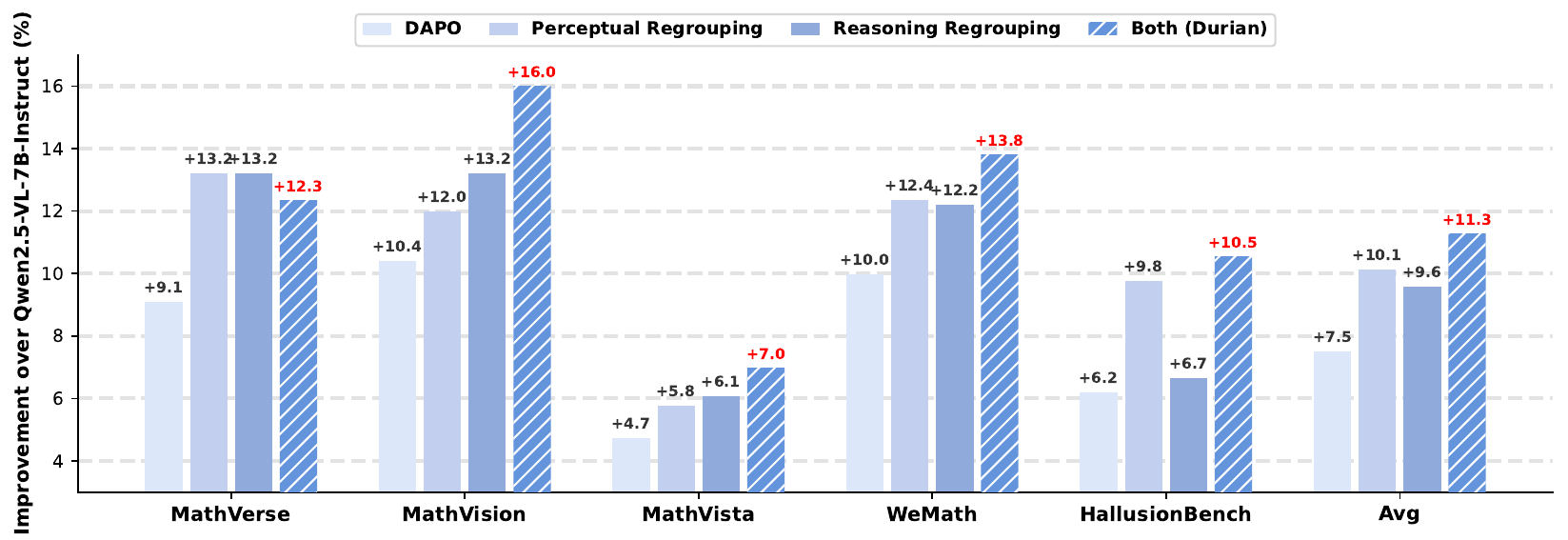}
    \caption{Acc Improvements of two re-grouping strategies over Qwen2.5-VL. We take DAPO as our backbone.}
    \label{fig:ablation}
\end{figure*}

To better understand the contribution of each component in Durian, we conduct ablation studies on five benchmarks, comparing four settings over Qwen2.5-vl: vanilla DAPO, DAPO with perceptual regrouping, DAPO with reasoning regrouping, and our Durian. Results are in Figure~\ref{fig:ablation}.

\textbf{The effects of Perceptual difficulty-based regrouping.}
Using perceptual difficulty-based regrouping alone yields consistent performance gains across benchmarks.
For instance, on HallusionBench, which is explicitly designed to evaluate perceptual ability, we observe an improvement of 3.4\% over vanilla DAPO.
This demonstrates that regrouping samples via spectral analysis of image patch covariances enhances the model’s perceptual grounding by mitigating the dominance of extremely easy or hard cases.

\textbf{The effects of Reasoning difficulty-based regrouping.}
The average accuracy under model confidence regrouping increases to 58.4, and it's notable to observe a 3.8\% gain on MathVerse, even surpassing the performance of our method, indicating that the model’s internal confidence estimation also serves as a reliable signal for stabilizing optimization.

The combination of both strategies achieves the best overall performance with an average accuracy of 59.3. 
This confirms that these two strategies provide complementary perspectives on samples, and their integration leads to more robust policy optimization.

\subsection{Hyper-parameter Sensitivity Analysis (RQ3)}

In this section, we analyze the effects of hyperparameters, including the number of perceptual, reasoning difficulty-based groups and $\alpha_{\text{Ori}}, \alpha_{\text{Percep}}, \alpha_{\text{Reason}}$.

\subsubsection{Groups under Perceptual difficulty-based strategy}
\begin{figure}[H]
    \vskip 0.1in
    \begin{center}
    \centerline{\includegraphics[width=\columnwidth]{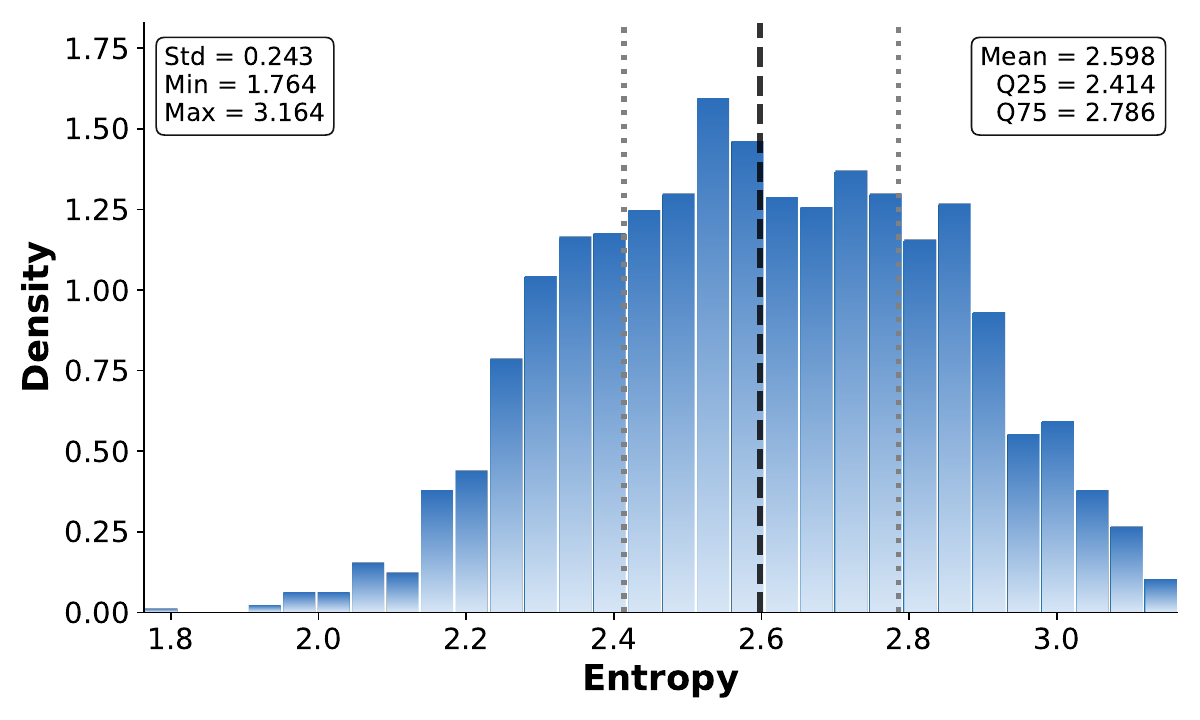}}
    \caption{The distribution of pre-calculated entropy on the Geometry3K. $x$ axis represents entropy, $y$ axis is the probability density, Q25 and Q75 denotes 25th and 75th percentiles, respectively  .}
    \vskip -0.36in
    \label{fig:entropy}
    \end{center}
\end{figure}

As shown in Figure \ref{fig:entropy} , to regroup samples by entropy, we adopt the 25th and 75th percentiles as thresholds. This quantile-based choice is inherently distribution-aware, as it adapts to the empirical spread of entropy values rather than relying on arbitrary fixed cutoffs. It produces a natural 1:2:1 partition of the data—approximately 25\% easy, 50\% medium, and 25\% hard—avoiding the issue of overly sparse or dense categories. Such a balance is desirable for stable optimization: each group contains sufficient samples to provide reliable intra-group statistical estimates, while extremely low- and high-entropy cases are isolated rather than allowed to dominate normalization. Moreover, this three-level categorization is semantically interpretable, with low entropy corresponding to simple scenes, high entropy to complex ones, and the middle range capturing moderately difficult cases (For further empirical analysis, see Appendix \ref{appendix:feasibility_of_2_proxy}). Detailed cases representing the entropy of these three categories are illustrated in Appendix \ref{sec:entropy_cases}.

\subsubsection{Groups under Reasoning difficulty-based strategy}
We investigate the impact of varying the number of groups in the reasoning-based regrouping strategy in Table \ref{tab:logp}. We observe that the performance is relatively stable across a wide range of groups, suggesting that our method is robust to this hyperparameter. For instance, on MathVista and HallusionBench, the accuracy is steadily improved as the number of groups increases to around 12–16, after which the results plateau with only minor fluctuations. This suggests that moderate group granularity is sufficient to capture meaningful variations in reasoning difficulty, while overly fine partitioning provides diminishing returns. A similar trend is observed on WeMath, where the performance peaks at 12 groups but remains competitive without significant degradation even when more groups are introduced.

\begin{table}[t]
\centering
\caption{The accuracy performance of different numbers of groups $b$ on five benchmarks.}
\label{tab:logp}
\resizebox{\columnwidth}{!}{
\begin{tabular}{c|ccccc}
\toprule
\textbf{Groups $b$} & \textbf{MathVerse} & \textbf{MathVision} & \textbf{MathVista} & \textbf{WeMath} & \textbf{HallusionBench} \\
\midrule
4  & 49.4 & 28.1 & 68.6 & 66.7 & 66.8 \\
6  & 50.6 & 27.1 & 69.7 & 70.2 & 70.0 \\
10 & 50.6 & 27.5 & 71.3 & 70.8 & 68.9 \\
12 & \textbf{52.3} & \textbf{28.5} & 70.9 & \textbf{71.4} & 69.6 \\
16 & 50.4 & 27.5 & \textbf{72.4} & 70.7 & 69.9 \\
20 & 51.5 & 28.1 & 71.3 & 70.4 & 68.8 \\
24 & 50.1 & 27.9 & 71.6 & 70.1 & 69.7 \\
32 & 50.5 & 26.9 & 71.6 & 68.7 & 70.7 \\
40 & 50.7 & 28.3 & 71.4 & 70.2 & 70.3 \\
48 & 50.0 & 27.7 & 71.5 & 69.5 & \textbf{70.9} \\
\bottomrule
\end{tabular}
}
\end{table}

\subsubsection{Analysis of Different Weighting Coefficients.}
\label{appendix:Coefficients_analysis}
We experiment with different combinations of three coefficients $\alpha_{Ori}$, $\alpha_{Percep}$, and $\alpha_{Reason}$ in Table \ref{tab:alpha_coefficients}. We can observe that while the performance on different benchmarks varies slightly with different settings, the method is relatively stable across a wide range of settings, with no significant degradation in results, indicating that our model is not overly sensitive to the specific choice of hyperparameters. This suggests that our method does not require extremely fine-tuned hyperparameters to perform effectively.

\begin{table}[htbp]
  \caption{The effects of different weighting coefficients (built upon DAPO) on 5 benchmarks}
  \label{tab:alpha_coefficients}
  \centering
  \resizebox{\columnwidth}{!}{
  \begin{tabular}{cccccccc}
    \toprule
    $\alpha_{\text{Ori}}$ & $\alpha_{\text{Percep}}$ & $\alpha_{\text{Reason}}$ & MathVerse & MathVision & MathVista & WeMath & HallusionBench \\
    \midrule
    0.1  & 0.2  & 0.7  & 50.7 & 28.6 & 71.6 & 71.8 & 71.2 \\
    0.15 & 0.25 & 0.6  & 50.8 & 29.0 & 71.6 & 70.6 & 70.8 \\
    0.2  & 0.1  & 0.7  & 51.2 & 28.4 & 71.5 & 70.4 & 71.0 \\
    0.3  & 0.1  & 0.6  & 51.7 & 27.8 & 70.7 & 70.6 & 70.8 \\
    0.4  & 0.3  & 0.3  & 51.9 & 27.9 & 71.4 & 70.5 & 71.1 \\
    0.6  & 0.2  & 0.2  & 50.4 & 28.8 & 72.2 & 71.0 & 71.4 \\
    0.7  & 0.1  & 0.2  & 51.1 & 28.3 & 70.3 & 71.1 & 69.8 \\
    \bottomrule
  \end{tabular}}
\end{table}

\section{Related Works}
In this section, we overview of the related studies. Specifically, we first discuss representative strategies to construct multimodal reasoning models, including chain-of-thought distillation, reinforcement learning, and visual tool integration. 
Then we introduce the RLVR and its optimization variants, and finally, we highlight the key differences between ours and existing approaches.

\subsection{Multimodal Reasoning Models}

\textbf{Chain-of-thought distillation.}
Supervised fine-tuning on long CoT data enables models to learn detailed reasoning traces, thereby improving reasoning accuracy. Specifically, building upon~\citep{multimodal_cot}, this strategy has proven effective through both transferring CoT-enhanced LLMs to multimodal settings~\citep{virgo} and training directly with multimodal reasoning data~\citep{llava}. Recent works explore different forms of intermediate reasoning supervision~\citep{instructblip, cosyn}.

\textbf{Reinforcement learning.}
Another line of research leverages RL to optimize reasoning trajectories beyond imitation. Most studies adopt PPO~\citep{ppo} or GRPO~\citep{DeepSeekMath}, with representative approaches such as~\citep{vlmr1, internvl3, llavar1} that apply RL across diverse domains.  We will elaborate on RLVR and GRPO in the following subsection (Section \ref{sec:rlvr}).

\textbf{Visual tool integration.} This paradigm moves beyond merely ``thinking about images'' toward actively querying, modifying, and generating visual information as intermediate steps in reasoning, forming a ``visual chain of thought''. 
The development of think-with-image can be roughly divided into three stages \citep{thinkwithimg}: from external tool exploration \citep{taco, vtriune}, through programmatic manipulation \citep{vipergpt, refocus}, to intrinsic imagination \citep{cotvla, blip3}. 
These three stages reflect interconnected capabilities---active exploration, structured reasoning, and generative planning---that together transform visual representations from static inputs into a dynamic workspace for thought. 

\subsection{Reinforcement Learning with Verifiable Reward}
\label{sec:rlvr}

RLVR~\citep{tulu3} is an optimization paradigm that replaces subjective reward scores with verifiable signals. Its core algorithm, GRPO~\citep{DeepSeekMath}, stabilizes training by comparing candidate responses within a group. Subsequent studies can be broadly categorized into two directions: data-centric methods, which expand the candidate and reward space through data manipulation or augmentation, and algorithm-centric methods, which refine GRPO to strengthen semantic grounding and coherent reasoning.

\textbf{Data-centric GRPO.}
This line of work enlarges the candidate set~\citep{mico} or restructures the training data~\citep{diff_prior, shuffler1} so that group comparisons capture richer behaviors. By manipulating data distributions~\citep{shuffler1} or augmenting inputs~\citep{visionmat, noisyrollout}, these methods expose models to a wider variety of responses, thereby increasing the likelihood of discovering high-quality verifiable signals.

\textbf{Algorithm-centric GRPO.}
In contrast, algorithm-centric methods refine how verifiable signals guide reasoning. Rather than expanding candidate sets, they adapt GRPO to enhance semantic grounding~\citep{perceptionr1, vis_ag_ft} and logical coherence~\citep{vision_r1, cold_start}. These approaches emphasize the role of visual grounding and promote reasoning chains where intermediate steps remain verifiable while supporting the final answer.

\textbf{Difference.} Compared with existing methods, we regroup the data in advantage calculation based on model response uncertainty and the entropy of images when computing \textit{std}, and sharing the \textit{std} within each group. This design prevents the model from overfitting to extreme samples and enhances its ability to capture the data distinction within each group.
\section{Conclusion}
In this work, we identify a critical challenge in GRPO-based reinforcement learning methods for multimodal reasoning tasks: \textbf{the \textit{std}-based group normalization is sensitive to extreme samples}, such as response groups that are almost entirely positive or negative. While this issue exists in GRPO in general, it is significantly amplified in multimodal settings due to the joint influence of perceptual complexity and reasoning uncertainty. 

To address this, we propose Durian, an effective \textbf{difficulty-aware re-grouping} strategy. By decomposing the difficulty into perceptual and reasoning aspects, we construct groups of samples with similar difficulty levels, allowing each group to share \textit{std} during normalization.
The normalized advantages with shared \textit{std} from both aspects are combined via element-wise combination, effectively integrating data complexity and model uncertainty while preserving intra-group distinctions. 
By applying over GRPO and DAPO, our strategy achieves 11.3\% average performance gains across multiple multimodal reasoning benchmarks.

\noindent \textbf{Limitations and Future Work.}
Durian inevitably introduces some hyperparameters, but we observe that performance across benchmarks remains relatively stable under a wide range of settings. This empirical robustness suggests that the Durian is not overly sensitive to precise hyperparameter tuning. Several directions remain open, such as more precise difficulty estimation and adaptive grouping strategies, which may better balance intra-group distinctions and mitigate extreme samples. Beyond technical refinements, the underlying principle of aligning optimization with sample difficulty also offers a general paradigm for stabilizing RL optimization with multimodal inputs.

\newpage
\section*{Impact Statements}
This paper presents work whose goal is to advance the field of machine learning. There are many potential societal consequences of our work, none of which we feel must be specifically highlighted here.

\bibliography{main}
\bibliographystyle{icml2026}

\appendix
\onecolumn
\section{Analysis of the occurrence of extreme samples, which \textit{std}-based group normalization is highly sensitive to.}
\label{appendix:analysis_motivation}
To further support the motivation of our paper and analyze the changes of extreme samples during training, we perform a detailed step-by-step analysis of reward statistics across 60 training steps with 512 samples and their 8 rollout rewards. The empirical evidence clearly shows that the existence of extreme samples is not an occasional event but a persistent and systemic phenomenon.

\begin{table}[htbp]
  \centering
  \caption{The statistics of rewards about extreme samples within the batch across 60 training steps.}
  \label{tab:extreme_samples_stats_rollout8}
  \resizebox{0.8\textwidth}{!}{%
  \begin{tabular}{lccccccc}
    \toprule
    Training steps & 1 & 10 & 20 & 30 & 40 & 50 & 60 \\
    \midrule
    Effective samples (participating in training) & 323 & 327 & 324 & 322 & 297 & 314 & 306 \\
    Extreme success (7 correct \& 1 wrong) & 41 & 39 & 48 & 66 & 78 & 60 & 82 \\
    Extreme failure (7 wrong \& 1 correct) & 78 & 89 & 74 & 51 & 54 & 54 & 51 \\
    Total Extreme Ratio & 36.8\% & 39.1\% & 37.7\% & 36.3\% & 44.4\% & 36.31\% & 43.5\% \\
    \bottomrule
  \end{tabular}}
\end{table}

First, groups with 8 identical rewards (i.e., variance = 0) constitute 35\%–46\% of all samples at every training step. We first exclude these groups for not participating in gradient updates.

Second, during the training process, there are \textbf{31\%–44\% samples exhibit the 7:1 extreme reward patterns} (i.e., 7/8 correct or wrong) among the remaining effective samples, which produce extremely small variance. Besides, the occurrence of this situation will increase as training deepens.

These findings demonstrate that \textbf{the instability of \textit{std}-based normalization is structural rather than incidental:} multimodal reasoning tasks naturally contain a large proportion of very easy and very hard samples, leading to unstable and unreliable advantage scaling. This directly motivates our difficulty-aware regrouping strategy, which stabilizes normalization by ensuring that variance is computed only within samples of comparable difficulty.

\section{Verify the feasibility of utilizing image entropy as a proxy for perceptual difficulty and model confidence as a proxy for reasoning difficulty.}
\label{appendix:feasibility_of_2_proxy}
\subsection{Image entropy as a proxy for perceptual difficulty.}
\textbf{Perceptual difficulty} in our framework is defined based on \textbf{the complexity of visual embeddings}, which we quantify using spectral analysis of image patch covariances. Specifically, the entropy of the eigenvalue distribution from the covariance matrix reflects the amount of variance across spatial features in the image. Researchers in prior works\citep{appendix_perceptual} support that: high entropy indicates a more diverse distribution of visual features, implying a richer and more complex visual structure. This complexity makes it more challenging for the visual model to recognize, and thus we associate higher entropy with greater perceptual difficulty.

\subsection{Model confidence as a proxy for reasoning difficulty.}
Researchers in \citep{appendix_reasoning1,appendix_reasoning2} propose that ``one measure of uncertainty is the predictive entropy of the output distribution, which measures the information one has about the output given the input[3]. The predictive entropy for an input sentence $\bf{x}$ is the conditional entropy ($H$) of the output random variable ($Y$) with realization $y$ given $\bf{x}$.''
\begin{equation}
\begin{aligned}
{\rm{PE}}({\bf{x}})=H(Y| {\bf{x}})=-\sum _{y}P(\,y| {\bf{x}})\mathrm{ln}P(\,y| {\bf{x}}).
\end{aligned}
\end{equation}
Researchers\citep{appendix_reasoning3} also hypothesize that when a model knows the answer to a particular question, it is confident in its response, and this would result in an answer distribution with small entropy. Conversely, when a model is unsure about its response, it will lead to an answer distribution with high entropy, thus implying a more challenging reasoning process.

This aligns directly with our formulation: the \textbf{sequence-level log probabilities} we compute are theoretically linked to the notion of \textit{semantic entropy} and represent the joint likelihood of the entire reasoning chain. A low log-probability corresponds to a flat or high-entropy output distribution, reflecting uncertainty in the reasoning trajectory, while a high log-probability corresponds to a confident, low-entropy distribution.

\subsection{Empirical validation.}
During the evaluation stage, we conduct an analysis focusing on the \textbf{questions that the model answered incorrectly} on two benchmarks. We want to examine whether these error samples are concentrated in the more difficult groups as defined by our difficulty metrics. The intuition behind this approach is that samples belonging to higher-difficulty groups—whether in terms of perceptual complexity or reasoning uncertainty—should naturally be harder for the model to tackle. Consequently, we expect these samples to exhibit higher error rates.

To achieve this, we use Gemini2.5 Pro to classify the sources of errors, distinguishing between \textbf{perceptual errors} and \textbf{reasoning errors}.

$\bullet$ For \textbf{perceptual errors}, we first group the images based on their visual entropy, then compute the proportion of incorrect answers within each group relative to the total number of perceptual errors.

$\bullet$ Similarly, for \textbf{reasoning errors}, we group the samples based on model confidence, and calculate the proportion of incorrect answers in each group relative to the total number of reasoning errors.

\begin{table}[H]
  \centering
  \caption{The error rate of perceptual difficulty groups in perceptual errors on two benchmarks.}
  \label{tab:perceptual_error_rate}
  \resizebox{0.5\textwidth}{!}{%
  \begin{tabular}{lccc}
    \toprule
        & low-entropy & medium-entropy & high-entropy \\
    \midrule
    Wemath  & 23.6\%   & 31.3\%      & 45.1\%       \\
    HallusionBench   & 21.2\%  & 29.6\%   & 49.2\%       \\
    \bottomrule
  \end{tabular}
  }
\end{table}

\begin{table}[H]
  \centering
  \caption{The error rate of reasoning difficulty groups in reasoning errors on two benchmarks.}
  \label{tab:reasoning_error_rate}
  \resizebox{0.95\textwidth}{!}{
    \begin{tabular}{lcccccccccc}
      \toprule
                            & group 1 (low confidence) & group 2 & group 3 & group 4 & group 5 & group 6 & group 7 & group 8 & group 9 & group 10 (high confidence) \\
      \midrule
      Wemath                & 13.4\%                   & 12.6\%  & 13.2\%  & 12.7\%  & 11.6\%  & 9.1\%   & 9.7\%   & 7.2\%   & 6.7\%   & 4.2\%                      \\
      HallusionBench        & 12.7\%                   & 11.9\%  & 11.2\%  & 11.7\%  & 10.0\%  & 9.8\%   & 9.0\%   & 8.2\%   & 8.3\%   & 7.1\%                      \\
      \bottomrule
    \end{tabular}
  }
\end{table}

As shown in Table \ref{tab:perceptual_error_rate} and Table \ref{tab:reasoning_error_rate}, the results align with our expectations: images with \textbf{low visual entropy} (indicating simplicity) correspond to \textbf{lower perceptual error rates}, and samples with \textbf{lower model confidence} (indicating greater uncertainty in the reasoning process) correspond to \textbf{higher reasoning error rates}. Our empirical findings are consistent with this intuition, further supporting the validity of our difficulty metrics.

\section{Experiment Settings}
\textbf{Reward Calculation.} We adopt a combination of format reward and accuracy reward as the final reinforcement learning signal. 
The two components are defined as follows:
\begin{equation}
r_{\text{format}} =
\begin{cases}
1, & \text{if the output format is correct}, \\
0, & \text{otherwise},
\end{cases}
\end{equation}

\begin{equation}
r_{\text{acc}} =
\begin{cases}
1, & \text{if the answer matches the ground truth}, \\
0, & \text{otherwise}.
\end{cases}
\end{equation}

The overall reward is computed as the weighted sum of the two:
\begin{equation}
r_{\text{overall}} = 0.1 \times r_{\text{format}} + 0.9 \times r_{\text{acc}}.
\end{equation}

A smaller weight is assigned to the format reward, since response formatting is relatively easy to learn compared with accuracy.

\section{Prompt Design}
We use a ``Thinking prompt'' to formalize the output of the model. It requires the model to put its reasoning process within \texttt{<think>...</think>} and the final answer in \verb|\boxed{}|. We keep the system prompt of Qwen2.5-VL \citep{qwen25} and prepend the ``Thinking prompt'' to the user message. The same format is used for both training and evaluation. The full instruction prompt is as follows:

\begin{tcolorbox}[colback=gray!10, colframe=black, title=Prompt Example]
\textbf{SYSTEM:}\\
You are a helpful assistant.\\
\textbf{USER:}\\
You FIRST think about the reasoning process as an internal monologue and then provide the final answer. The reasoning process MUST BE enclosed within \texttt{<think>...</think>} tags. The final answer MUST BE put in \verb|\boxed{}|.\texttt{<QUESTION>}
\end{tcolorbox}

\section{Analysis of the effect of rollout size on performance and stability.}
\label{sec:rollout_size}
\begin{table}[htbp]
  \centering
  \caption{The effects of rollout size (built upon DAPO) on 5 benchmarks.}
  \label{tab:group_size_effects}
  \resizebox{0.6\textwidth}{!}{%
  \begin{tabular}{cccccc}
    \toprule
    rollout & MathVerse & \textbf{MathVision} & MathVista & WeMath & HallusionBench \\
    \midrule
    2       & 48.7      & 27.1                & 70.1      & 69.7   & 67.3           \\
    4       & 50.1      & 28.4                & 71.5      & 70.2   & 68.9           \\
    8       & 51.9      & 29.0                & 72.2      & 71.8   & 71.4           \\
    16      & 52.1      & 29.2                & 72.1      & 71.2   & 71.2           \\
    24      & 51.7      & 29.0                & 71.9      & 71.9   & 71.5           \\
    32      & 51.9      & 28.9                & 72.2      & 71.0   & 71.3           \\
    \bottomrule
  \end{tabular}}
\end{table}

We observed from Table \ref{tab:group_size_effects} that when the rollout size is smaller than 8, the performance improves as the rollout size increases. Notably, when the number of rollouts is reduced to 2, the model reverts to PPO. As the rollout size continues to increase beyond 8, the improvement in performance becomes less pronounced, eventually stabilizing at a stable value.

These results indicate that while increasing the number of rollouts can lead to better performance, after a certain point, beyond which further increases in group size do not significantly contribute to performance improvement. This shows the importance of selecting an appropriate group size to balance computational cost and model performance.

\section{The Use of Large Language Models(LLMs)}
We conducted a study on improving GRPO to further enhance the reasoning capability of MLLMs, achieving substantial performance gains on datasets such as MathVerse \citep{mathverse}, MathVision \citep{mathvision}, and MathVista \citep{mathvista}. During the preparation of this manuscript, we used LLMs to assist with tasks such as grammar correction, language refinement, and logical checking. However, we confirm that no outputs from the LLMs were directly used; instead, all content underwent careful verification and reconstruction by the authors.

\section{Case Study}
\subsection{Perceptual difficulty-based re-grouping cases}
\label{sec:entropy_cases}
\begin{figure}[H]
    \centering
    \includegraphics[width=0.9\textwidth]{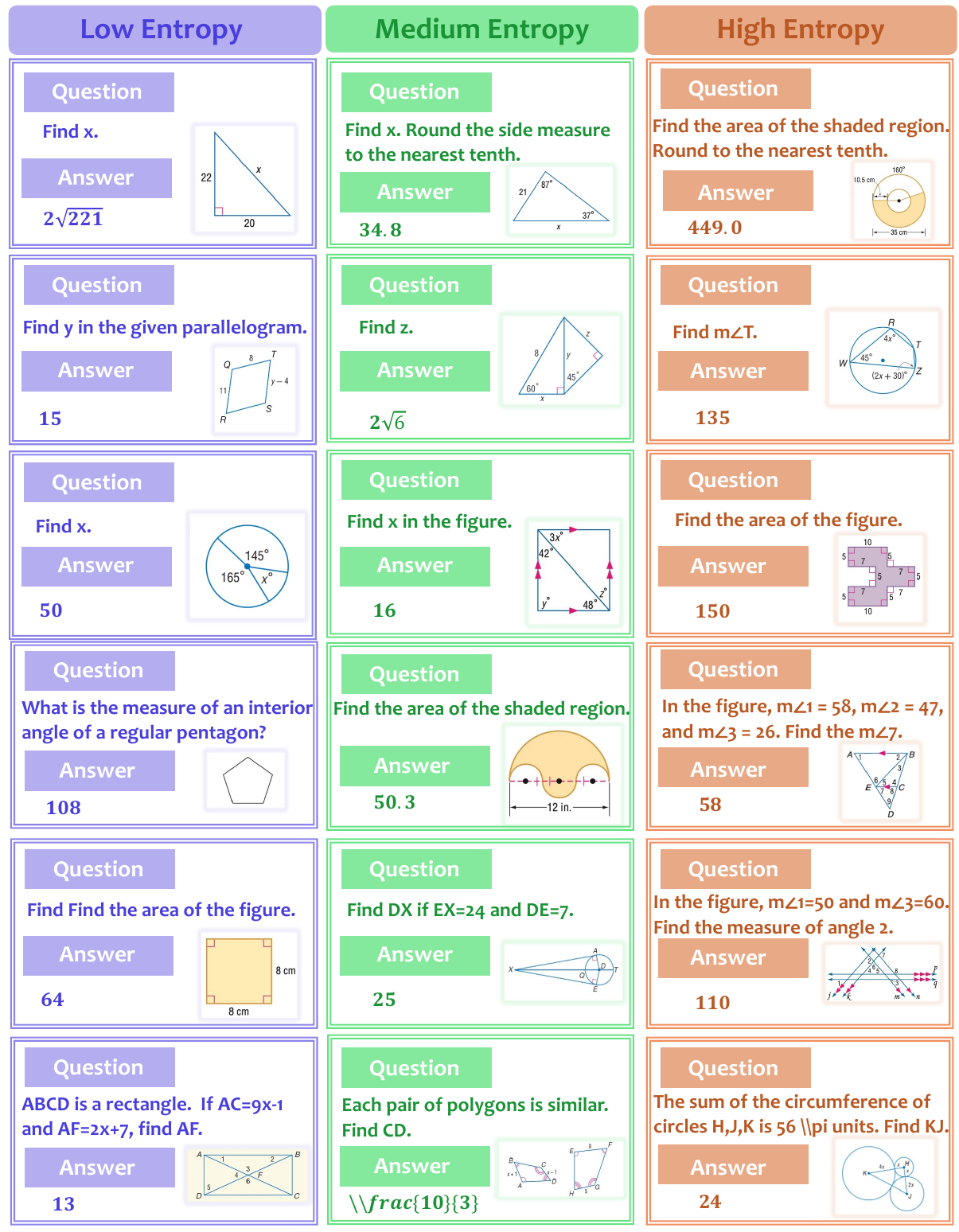}
    \caption{Illustrative examples of different levels of entropy}
    \label{fig:diffculty_level}
\end{figure}

\subsection{Demonstration of improved perception and reasoning capabilities}
\begin{figure}[H]
    \centering
    \includegraphics[width=0.9\textwidth]{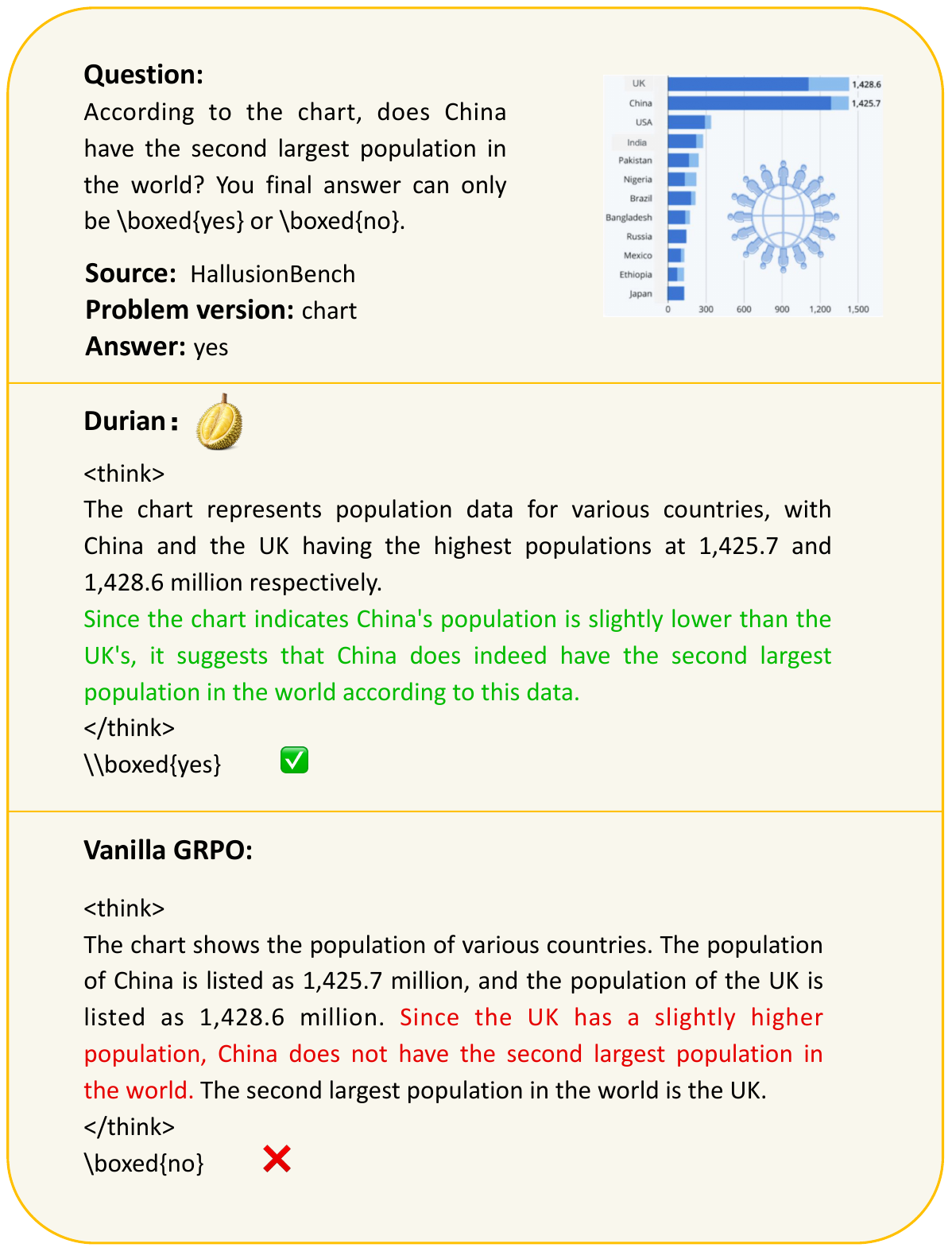}
    \caption{Case Study 1 showing improved reasoning capability on HallusionBench over vanilla GRPO.}
    \label{fig:hallusion}
\end{figure}

\begin{figure}[H]
    \centering
    \includegraphics[width=0.9\textwidth]{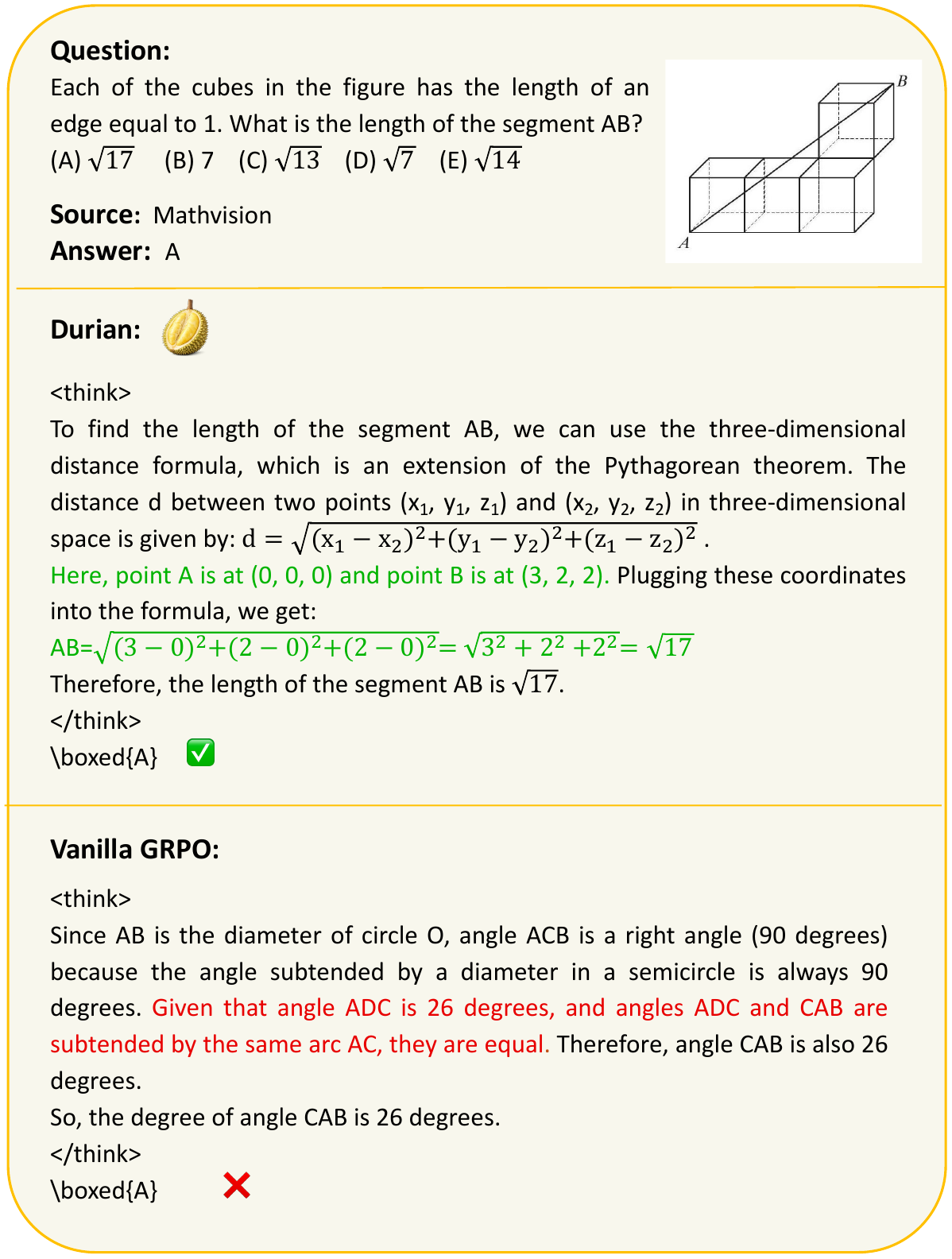}
    \caption{Case Study 2 showing improved reasoning capability on Mathvision over vanilla GRPO.}
    \label{fig:mathvision}
\end{figure}

\begin{figure}[H]
    \centering
    \includegraphics[width=0.9\textwidth]{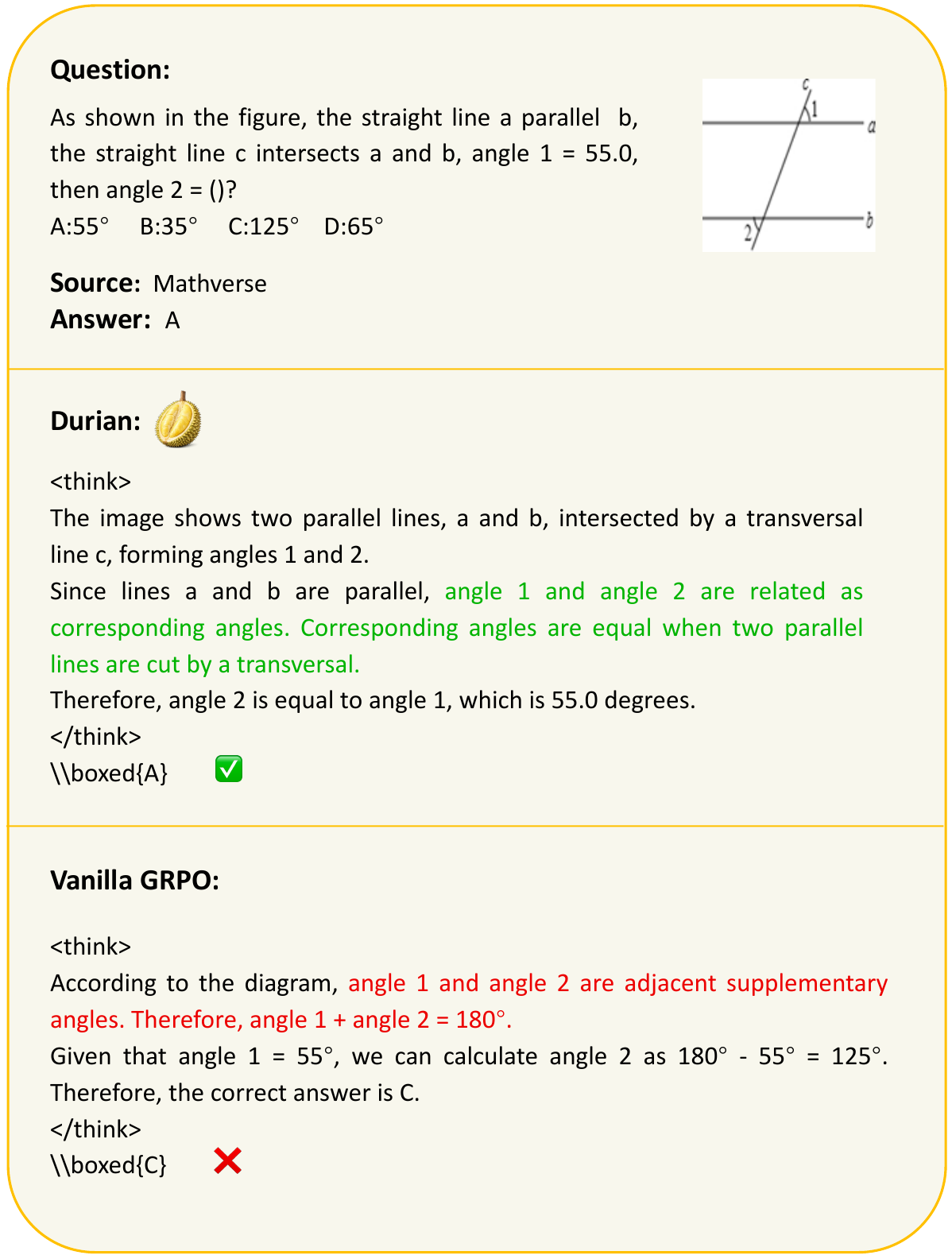}
    \caption{Case Study 2 showing improved reasoning capability on Mathverse over vanilla GRPO.}
    \label{fig:mathverse}
\end{figure}

\begin{figure}[H]
    \centering
    \includegraphics[width=0.9\textwidth]{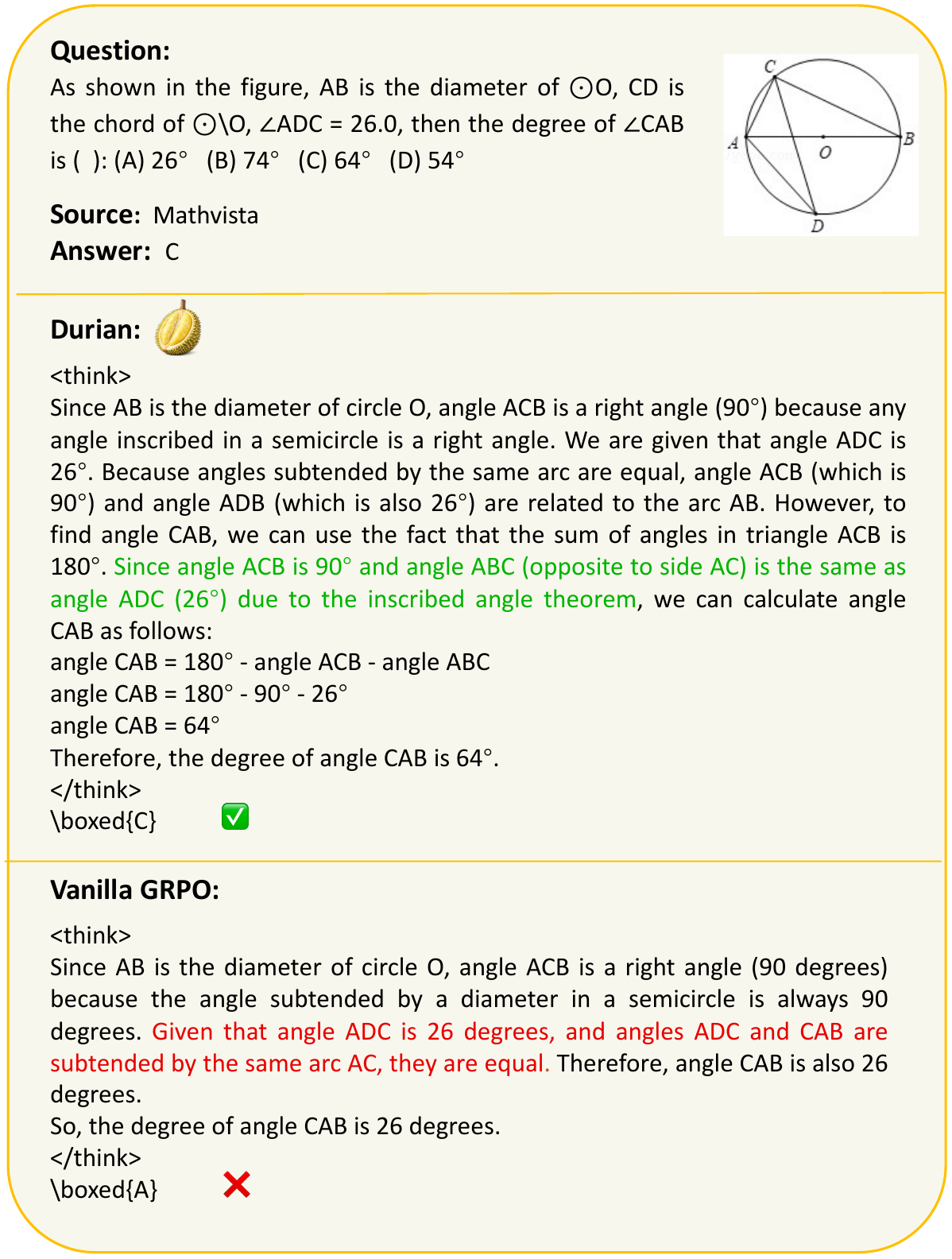}
    \caption{Case Study 3 showing improved reasoning capability on Mathvista over vanilla GRPO.}
    \label{fig:mathvista}
\end{figure}

\begin{figure}[H]
    \centering
    \includegraphics[width=0.9\textwidth]{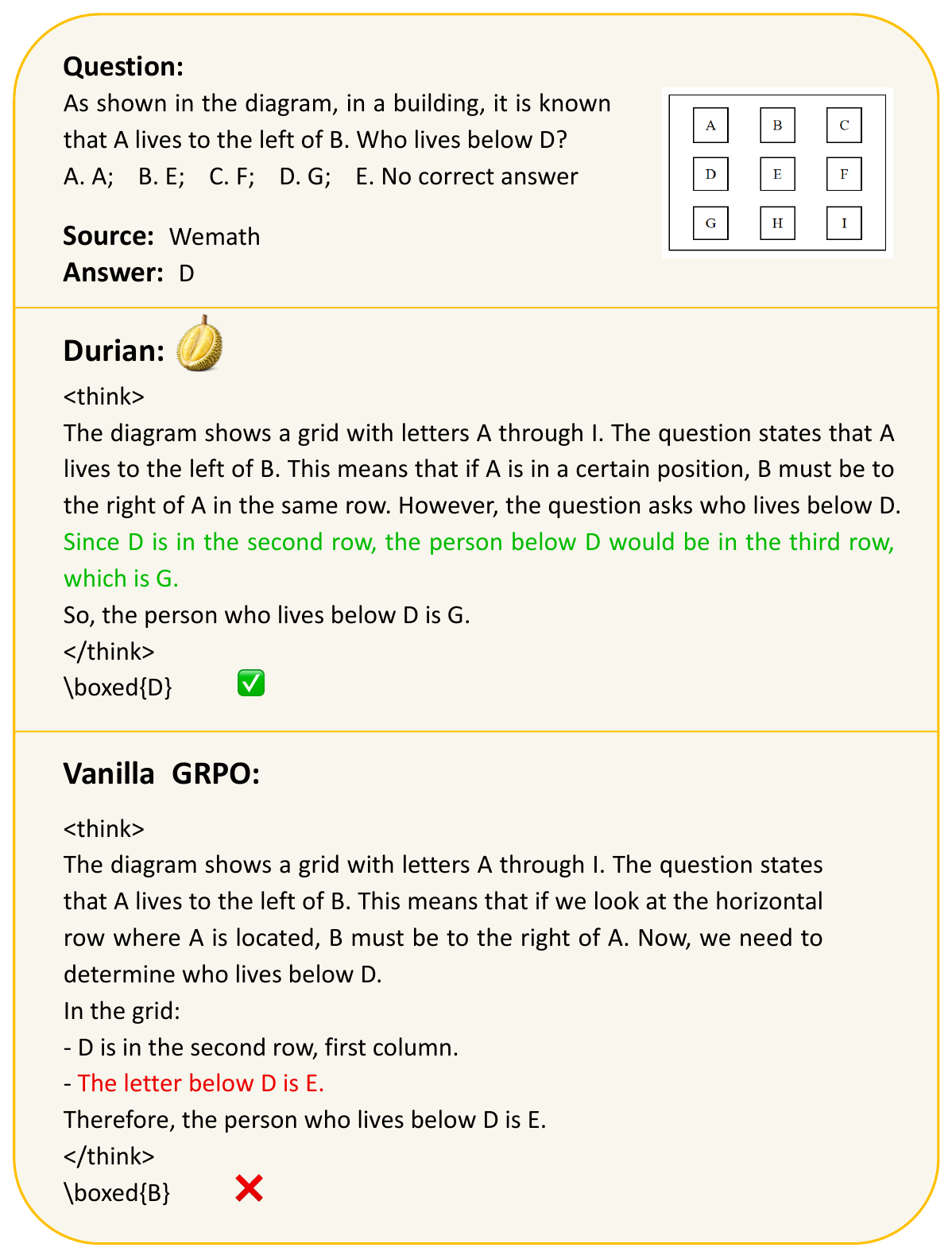}
    \caption{Case Study 4 showing improved reasoning capability on Wemath over vanilla GRPO.}
    \label{fig:wemath}
\end{figure}

\end{document}